\documentclass[letterpaper, 10 pt, conference]{ieeeconf}  

\IEEEoverridecommandlockouts                              

\usepackage{amsmath} 
\usepackage{amssymb}  
\usepackage{subcaption}
\usepackage{xcolor}
\usepackage{graphicx}
\usepackage{multirow}
\usepackage[colorlinks = true,
            urlcolor = black, 
            linkcolor = black]{hyperref}
\usepackage[overlay, absolute]{textpos}

\DeclareMathOperator*{\argmax}{arg\,max}

\title{\LARGE \bf
End-to-End Reinforcement Learning for Torque Based \\ Variable Height Hopping
}

\author{Raghav Soni$^{1,*}$, Daniel Harnack$^{2,*}$, Hannah Isermann$^{2,\dagger}$, Sotaro Fushimi$^{3}$, Shivesh Kumar$^{2}$, Frank Kirchner$^{2}$
\thanks{This work was supported by the VeryHuman Project funded by the German Aerospace Center (DLR) with federal funds (FKZ: 01IW20004) and is additionally supported with project funds from the federal state of Bremen for setting up the Underactuated Robotics Lab (Grant Number: 201-342-04-2/2021-4-1)}
\thanks{$^{1}$Department of Electronics Engineering, Indian Institute of Technology (Banaras Hindu University), Varanasi, India.}%
\thanks{$^{2}$DFKI GmbH Robotics Innovation Center, Bremen, Germany}%
\thanks{$^{3}$Undergraduate Course Program of Mechanical and Systems Engineering, Kyoto University, Kyoto, Japan}
\thanks{$^{\dagger}$Corresponding author: {\tt\small hannah.isermann@dfki.de.}}
\thanks{$^*$Both authors contributed equally}
}

\begin{document}
\begin{textblock*}{\textwidth}(0.75in, 0.25in)
\noindent \footnotesize{\copyright~2023 IEEE.  Personal use of this material is permitted.  Permission from IEEE must be obtained for all other uses, in any current or future media, including reprinting/republishing this material for advertising or promotional purposes, creating new collective works, for resale or redistribution to servers or lists, or reuse of any copyrighted component of this work in other works.

Cite as: \href{https://doi.org/10.1109/IROS55552.2023.10342187}{R. Soni, D. Harnack, H. Isermann, S. Fushimi, S. Kumar and F. Kirchner, "End-to-End Reinforcement Learning for Torque Based Variable Height Hopping," \textit{2023 IEEE/RSJ International Conference on Intelligent Robots and Systems (IROS)}, Detroit, MI, USA, 2023, pp. 7531-7538, doi: 
10.1109/IROS55552.2023.10342187}.
}
\end{textblock*}

\maketitle
\thispagestyle{empty}
\pagestyle{empty}

\begin{abstract}
Legged locomotion is arguably the most suited and versatile mode to deal 
with natural or unstructured terrains. Intensive research into dynamic walking and 
running controllers has recently yielded great advances, both in the optimal 
control and reinforcement learning (RL) literature. 
Hopping is a challenging dynamic task involving a flight phase and has the potential to increase the traversability of legged robots. 
Model based control for hopping typically 
relies on accurate detection of different jump phases, such as lift-off or 
touch down, and using different controllers for each phase. In this paper, 
we present a end-to-end RL based torque controller that learns to implicitly detect 
the relevant jump phases, removing the need to provide manual heuristics for 
state detection. We also extend a method for simulation to reality transfer 
of the learned controller to contact rich dynamic tasks, resulting in successful 
deployment on the robot after training without parameter tuning.
\end{abstract}


\section{Introduction}
\label{sec_intro_related_works}
Dynamic legged locomotion evolved as a versatile strategy to traverse
natural or unstructured terrains.
Thus, legged robots such as 
quadrupeds and humanoids are popular for applications performed 
in these environments, either autonomously or alongside a human. 
Quasi-instantaneously making and breaking contacts with the environment 
is an integral part of legged locomotion, which leads to highly nonlinear, non-smooth dynamics.
Thus, from a control perspective, dynamic legged locomotion 
requires significantly more complex algorithms than e.g. wheeled locomotion.  
Whereas the problem of dynamic walking on real robots has been solved by various techniques from 
optimal control (OC)~\cite{mombaur2008optimal, osumi2006time, hereid20163d} or 
reinforcement learning (RL)~\cite{chen2022learning, hwangbo2019learning, li2021reinforcement, jain2019hierarchical, xie2020learning, margolis2022rapid}, there is considerably 
less research for the even more dynamic locomotion type of hopping. 
Hopping can increase a system's mobility,
since it allows for leaping over obstacles that cannot be surpassed otherwise~\cite{hawkes2022engineered, bellegarda2020robust}. 
However, hopping incurs even more control complexity since there is a considerable flight phase during which the system has limited possibilities to adjust for the impact, and the center of mass trajectory is largely determined when the feet leave the ground.

\begin{figure}[!htbp]
	\centering
	\includegraphics[width=\linewidth]{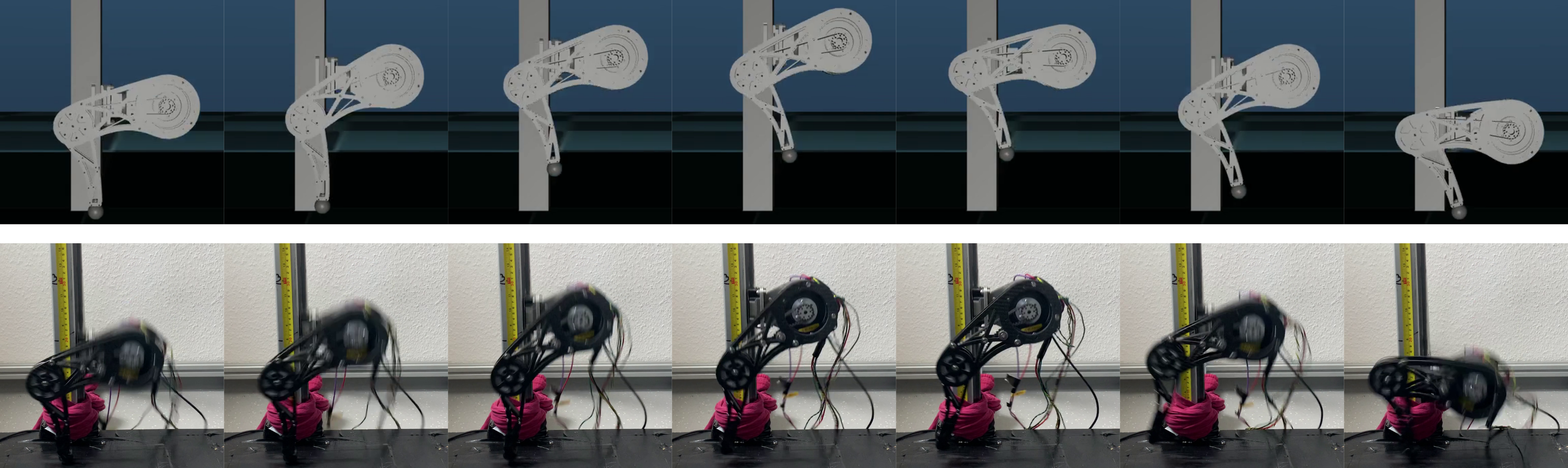}  
	\caption{RL based torque controlled jumping snapshots in simulation and on the 
		real robot.
	}
	\label{fig:jumping}
\end{figure}

\begin{figure}[!htbp]
	\centering
	\includegraphics[width=\linewidth]{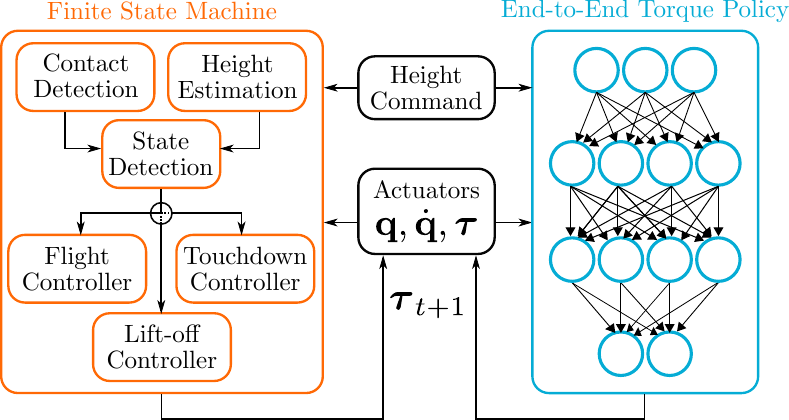}  
	\caption{Comparison of control concepts.
	}
	\label{fig:control_concepts}
\end{figure}

The canonical system to study hopping, which is also used in this paper, 
is a single hopping leg or monoped. Indeed, one of the earliest robotic systems showing dynamic 
legged locomotion was a single  
leg that could navigate a flat surface by jumping~\cite{raibert1984experiments}. 
To control the hopping height, a heuristically tuned force controller was used, motivated by an energy shaping algorithm~\cite{raibert1984hopping}. These seminal 
studies sparked a wealth of research into the control 
of single legged hopping machines.
A common theme of such controllers is the 
reliance on detection of various states, e.g. lift-off, peak attitude, touchdown, and minimum attitude 
\cite{raibert1984hopping, zeglin1991uniroo, harbick2002controlling, lebaudy1993control, zhao2020bio, ramos2020hoppy}. 
The full jumping controller is then realized as a state machine, 
where PD controllers are typically used during flight phases, while stance phase states 
are directly controlled by torque or force. 
Deploying these controllers on hardware requires hand tuning parameters and system specific 
adaptations to account for model inaccuracies or unmodelled dynamics. Also, the detection of different 
jump states and appropriate control output during the lift-off phase relies on 
accurate height estimation and 
contact detection, for which further heuristics are typically employed. 

RL offers the promise to alleviate these issues. 
We hypothesize that, since neural networks are universal function approximators 
\cite{pinkus1999approximation, lu2017expressive}, learning based controllers with neural network function approximators are able to implicitly detect relevant jump phases and thus realize a unified hopping controller without the need for explicit state transition heuristics. Even 
more, a fully integrated end-to-end solution for hopping should be possible via RL, mapping 
only proprioceptive feedback, i.e. actuator positions and velocities, 
to direct torque control, since this proprioceptive data is theoretically
sufficient to implement a variable jumping height controller. The comparison 
of our method to a classical approach is visualized in Fig.~\ref{fig:control_concepts}.

Whereas several previous studies utilized learning based controllers for jumping
\cite{bellegarda2020robust, kuang2018learning}, they focussed on single 
leaps, making implicit detection of jump phases less critical than for 
continuous jumping. From the data, it can also not be derived whether such a 
phase detection is actually realized. 
Recently, a RL based continuous hopping controller with adjustable 
jumping height for a small quadruped was developed in~\cite{bogdanovic2021model}. While 
this shows the feasibility of a unified controller that does not require a state machine, it still relied on height estimation and PD control, and thus can not be considered as a truly end-to-end learning approach. The widespread use of PD controllers in RL research is in part a consequence of the low sample complexity of many algorithms, which necessitates training in 
simulation. Using PD controllers reduces the requirements on the 
accuracy of the dynamics simulation and thus increases the chances of successful simulation to 
reality transfer. However, this requires tuning of PD gains for a successful sim2real transfer and may additionally hinder performance in highly dynamic tasks such as jumping, where direct torque control can unlock the full dynamical capabilities of a system~\cite{chen2022learning, wolpert2011principles}. 

In summary, all previous approaches from classical control and RL require a subset 
of height estimation, contact detection, hyperparameter tuning, PD control, or a behavior state machine.
In this paper, we present a RL based method that requires none of the above.
We show successful training and simulation to reality transfer of a torque controller with implicit jumping phase detection and controllable jumping height, while only relying on proprioceptive 
feedback~\footnote{\url{https://github.com/dfki-ric-underactuated-lab/hopping_leg}}. To achieve this, we draw inspiration from energy shaping for the design of the 
reward function, and extend a previous technique for accurate simulation to reality transfer~\cite{kaspar2020sim2real} to higher dimensional parameter spaces and dynamic, contact rich tasks. 
To the best knowledge of the authors, such a controller is described for the first time for a monoped.

\begin{figure}[!htbp]
	\begin{subfigure}{.47\columnwidth}
		\centering
		\includegraphics[width=\linewidth]{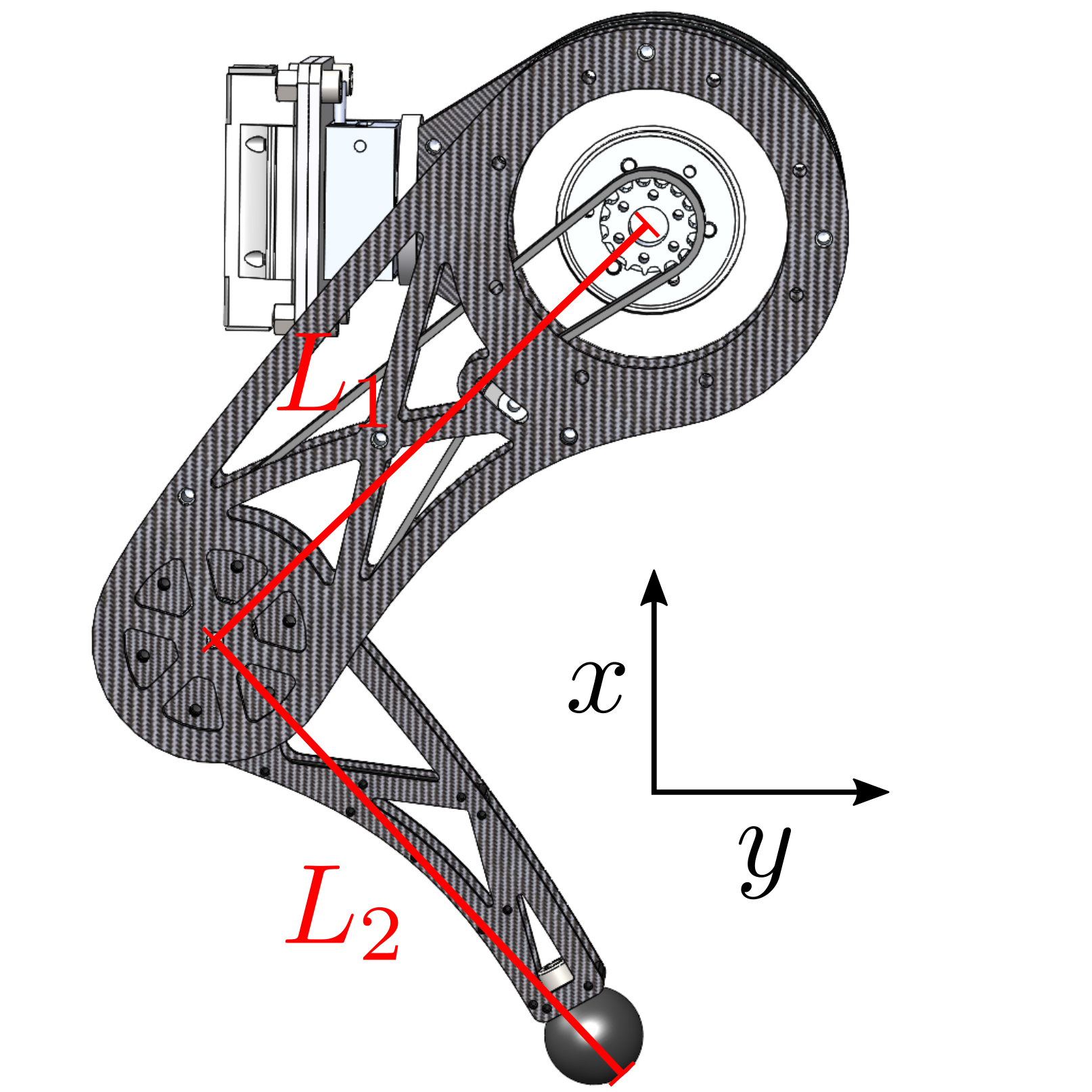}  
	\end{subfigure}
	\hfill
	\begin{subfigure}{.47\columnwidth}
		\centering
		\includegraphics[width=\linewidth]{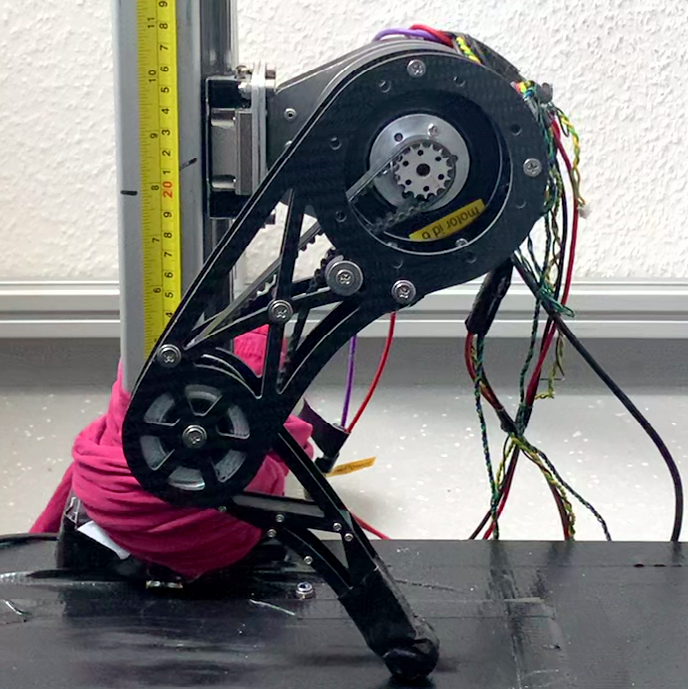}  
	\end{subfigure}
	\caption{Hopping leg used in the experiments.
	}
	\label{fig:robot}
\end{figure}

\section{Materials and Methods}
\label{sec:mats_and_meths}
\subsection{Robotic System}
The robot used for the experiment is a custom-made 3 degrees of freedom (DOF) hopping leg system, mounted on a vertical rail with 1 passive DOF and 2 active DOFs. 
Fig.~\ref{fig:robot} shows a photo of the system, along with a 3D design model. The 2 active DOFs in the leg are actuated via quasi-direct drive motors qdd100 from mjbots~\cite{mjbots} operating at a frequency of $200Hz$. While the shoulder joint shares the joint axis with the motor axis, the elbow joint is
driven by a motor placed at the shoulder via a belt drive with transmission ratio = 1:2. The housing is a light weight carbon fiber construction.

\subsection{Energy Shaping}
As a baseline comparison, we implemented a classical energy shaping (ES) controller in simulation~\cite{Tedrake2023}. 
This ES controller is part of a finite state machine~\cite{hyon2003dynamics}. As shown in Fig.~\ref{fig:state_machine} the state machine consists of three states:  
\begin{itemize}
	\item \textit{Lift-off}: This phase is used to apply the desired energy with the ES controller for the next jump. It ends when the leg loses its ground contact.
	\item \textit{Flight}: In the flight phase, the leg prepares for the touchdown by moving into a predefined pose.  The phase ends with the first contact of the leg with the ground.
	\item \textit{Touchdown}: During touchdown, the leg damps its movement using high damping and low positional gains. It ends when the base velocity $\dot x \geq 0$.
\end{itemize}
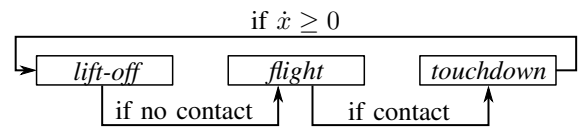
\begin{figure}[b]
	\centering
\begingroup%
  \makeatletter%
  \providecommand\color[2][]{%
    \errmessage{(Inkscape) Color is used for the text in Inkscape, but the package 'color.sty' is not loaded}%
    \renewcommand\color[2][]{}%
  }%
  \providecommand\transparent[1]{%
    \errmessage{(Inkscape) Transparency is used (non-zero) for the text in Inkscape, but the package 'transparent.sty' is not loaded}%
    \renewcommand\transparent[1]{}%
  }%
  \providecommand\rotatebox[2]{#2}%
  \newcommand*\fsize{\dimexpr\f@size pt\relax}%
  \newcommand*\lineheight[1]{\fontsize{\fsize}{#1\fsize}\selectfont}%
  \ifx\svgwidth\undefined%
    \setlength{\unitlength}{211.9529947bp}%
    \ifx\svgscale\undefined%
      \relax%
    \else%
      \setlength{\unitlength}{\unitlength * \real{\svgscale}}%
    \fi%
  \else%
    \setlength{\unitlength}{\svgwidth}%
  \fi%
  \global\let\svgwidth\undefined%
  \global\let\svgscale\undefined%
  \makeatother%
  \begin{picture}(1,0.20523197)%
    \lineheight{1}%
    \setlength\tabcolsep{0pt}%
    \put(0,0){\includegraphics[width=\unitlength,page=1]{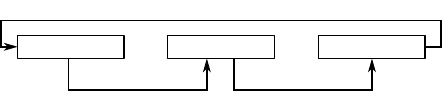}}%
    \put(0.17901981,0.01225502){\color[rgb]{0,0,0}\makebox(0,0)[lt]{\lineheight{1.25}\smash{\begin{tabular}[t]{l}if no contact\end{tabular}}}}%
    \put(0.5929533,0.01141267){\color[rgb]{0,0,0}\makebox(0,0)[lt]{\lineheight{1.25}\smash{\begin{tabular}[t]{l}if contact\end{tabular}}}}%
    \put(0.42029796,0.17196995){\color[rgb]{0,0,0}\makebox(0,0)[lt]{\lineheight{1.25}\smash{\begin{tabular}[t]{l}if $\dot x \geq 0$\end{tabular}}}}%
    \put(0.45223842,0.08303264){\color[rgb]{0,0,0}\makebox(0,0)[lt]{\lineheight{1.25}\smash{\begin{tabular}[t]{l}\textit{flight}\end{tabular}}}}%
    \put(0.10442009,0.0828106){\color[rgb]{0,0,0}\makebox(0,0)[lt]{\lineheight{1.25}\smash{\begin{tabular}[t]{l}\textit{lift-off}\end{tabular}}}}%
    \put(0.74060536,0.08329476){\color[rgb]{0,0,0}\makebox(0,0)[lt]{\lineheight{1.25}\smash{\begin{tabular}[t]{l}\textit{touchdown}\end{tabular}}}}%
  \end{picture}%
\endgroup%

	\caption{State machine used for the energy shaping controller.}
	\label{fig:state_machine}
\end{figure}

\subsubsection{Controller design}
As mentioned above, the desired energy $E_d$ which is required to jump to a desired base height $x_d$ hast to be applied during the lift-off phase.   
For simplicity, we assume the robot to be a point mass $m$ at its base. In this case, the required energy can be calculated with:
\begin{equation}
E_d = mgx_d
\end{equation}
Here, $g$ is the gravitational acceleration. 

Accordingly, we can estimate the reached energy $E_{j-1}$ from the last jump using the estimated jumping height $x_{j-1}$:
\begin{equation}
E_{j-1} = mgx_{j-1}
\end{equation}
For reaching this energy, we needed to apply a feed-forward force $F_{f, j-1}$ to the ground while the leg had ground contact. For the next jump, we estimate the new feed-forward force with:
\begin{equation}
F_{f,j} = \frac{mg\left(x_d-x_{0,j}\right)}{\Delta x_{l,j}}
\end{equation}
Here, $x_{0,j}$ is the current minimum base height after the touchdown, and $\Delta x_{l,j}$ is the the expected distance to be covered during the current lift-off, calculated from $x_{0,j}$.  
Since the formerly applied force is proportional to the reached energy $E_{j-1}$ we can write:
\begin{equation}
E \propto k F_{f}
\end{equation}
and as the feed-forward force is almost constant:
\begin{equation}
\dot{E} \propto \dot k F_{f}
\end{equation}
Thus, we can control the energy in the system by altering the gain $k$. To control the gain $k$ the following update rule has been used:
\begin{equation}
k_{j} = k_{j-1}\left(\frac{E_d}{E_{j-1}}\right)^2
\end{equation}
As joint torque controller, Cartesian stiffness control with the following control law was used~\cite{diCarlo2020software}:
\begin{equation}
\label{eq:carstesian_stiffness}
\tau = J^T(q) \left[k_j \begin{pmatrix}F_{f,j}\\0\end{pmatrix} + k_{p,y} \begin{pmatrix}0\\-y\end{pmatrix} + k_{d,y} \begin{pmatrix}0\\-\dot y\end{pmatrix} \right]
\end{equation}
Here, $\tau$ is the vector of desired joint torques, $J^T(q)$ is the transpose of the hybrid Jacobian at the end-effector and $K_{p,y}$ and $K_{d,y}$ are the Cartesian PD gains in $y$ direction (refer to Fig.~\ref{fig:robot}).
As shown in (\ref{eq:carstesian_stiffness}), the PD terms of the Cartesian stiffness controller are only responsible to maintain the $y$ position of the end-effector, while the energy shaping control is used to apply the forces in $x$ direction.

\subsubsection{Jumping height estimation}
For the ES controller, a height feedback is necessary. Therefore, a proprioceptive height estimation has been implemented. During flight phase, no additional forces can be applied to the system. Hence, we can expect the base acceleration $\ddot x$ to be roughly equal to the gravitational acceleration $g$. Due to the proprioceptive feedback, we know the lift-off position $x_l$ and velocity $\dot x_l$. Thus, the current base height $x$ during flight phase can be calculated with:
\begin{equation}
	x = \frac{1}{2} g \left(t^2-t_l^2\right) - g\, t_l \left(t-t_l\right) + \dot x_l \left(t-t_l\right) + x_l
\end{equation}
Here, $t$ is the current time and $t_l$ is the time at lift-off.
During stance phase, the current height is calculated from forward kinematics.

\subsubsection{Simulation and real system parameters}
The simulations of the ES controller have been performed using PyBullet physics simulation~\cite{pybullet}. For the simulation a control frequency of $400Hz$ has been used to control the joint torques. The parameters 
$k_{j=0} = 1.0, k_{p,y} = 10.0, k_{d,y} = 3.0$ were optimized manually. For the real system, the control frequency has been reduced to $200$ Hz.

\subsection{Reinforcement Learning}

\subsubsection{Problem Formulation}

The hopping leg problem is formulated as a 
Markov Decision Process (MDP), where the agent, i.e. the controller in this case, interacts with the environment, i.e. the leg and its surroundings. 
A MDP is given by a tuple $(\mathcal{S}, \mathcal{A}, \mathcal{P}, \mathcal{R})$, where $\mathcal{S}$ is the set of states called the state space, $\mathcal{A}$ is the set of actions called the action space, $\mathcal{P}(s_{t+1} \mid s_t, a_t)$ the probability that taking action $a_t$ in state $s_t$ will lead to state $s_{t+1}$, and $\mathcal{R}(s_{t}, a_{t}, s_{t+1})$ the expected immediate reward for transitioning from $s_t$ to  $s_{t+1}$ by taking the action $a_{t}$. 
At each time step t, an action $a_t \sim \pi(a_t \mid s_t)$ is sampled from the policy given the current state $s_t$. The objective of RL is to optimize the policy $\pi$ such that the expected return is maximized. 

From the variety of RL algorithms, we choose Soft Actor-Critic (SAC) \cite{https://doi.org/10.48550/arxiv.1801.01290}, a state-of-the-art off-policy algorithm, since it is relatively sample-efficient, stable, and requires little to no hyperparameter tuning. SAC aims to maximize the expected reward while also
maximizing the policy entropy $\mathrm{H}$. The objective is formulated as
\begin{equation*}
\resizebox{0.97\hsize}{!}{$
	\pi^* = \argmax_{\pi} \: \underset{a \sim \pi}{\mathrm{E}} \left[ \sum_{t=0}^{\infty} \gamma^t \biggl ( \mathrm{R}(s_{t}, a_{t}, s_{t+1}) + \alpha \mathrm{H}(\pi( \cdot \mid s_t)) \biggr ) \right].
$}
\end{equation*}
Maximizing the entropy as a secondary objective leads to policies that 
are maximally variable while performing the task, making them intrinsically robust.

For continuous actions, exploration is commonly done in action space. At each time step, a noise vector $\epsilon_t$ is sampled from a Gaussian distribution and added to the action output, such that $\pi(a_t \mid s_t) \sim \mu(s_t , \theta_\mu) + \mathcal{N}(0, \sigma^2)$, where $\mu$ is the deterministic policy and $\theta_\mu$
its parameters.
We use the modification of generalized state dependent exploration (gSDE) \cite{raffin2020generalized}. 
Here, the noise vector is a function of the state and the policy features $z_\mu(s_t, \theta_{z_\mu})$, which is the last layer before the deterministic output $\mu(s_t) = \theta_\mu z_\mu(s_t, \theta_{z_\mu})$, i.e. $\epsilon_t(s_t, \theta_\epsilon) = \theta_\epsilon z_\mu(s_t)$.
With gSDE, the action for a given state $s_t$ remains the same until the noise parameters are sampled again. This promotes more consistent exploration and results in reduced shaky behavior on hardware~\cite{raffin2020generalized}.

\subsubsection{Network architecture}
The policy is modeled with a multilayer perceptron (MLP) with four hidden layers of 256, 256, 128, and 128 neurons. The activation function is ReLU. The critic network is modeled by a separate network with the same architecture. LSTM policy networks were also tried but offered no empirical advantage. The policy is inferred at the operating frequency of $200Hz$.

\subsubsection{Observation and action space}
The hopping leg system has no additional sensors apart from the joint encoders. 
Thus, only normalized joint positions and velocities, and the desired jumping height over the three last time-steps $t$, $t-1$, and $t-2$ constitute the observation state. Joint data over multiple time-steps is empirically found to be essential to produce the desired behaviour with implicit contact detection. Hence, the observation space is $s \: \epsilon \: \mathbb{R}^{3 \times 5 = 15}$. 
The action space consists of the normalized output motor torques, which are later scaled up before being sent as the torque commands. The action space is thus $a \: \epsilon \: \mathbb{R}^{2}$.

\subsubsection{Reward}
The total reward at each time step is a weighted sum of positive gains and negative penalties, encoding behaviours to be encouraged or precluded. 
The reward comprises the following components:
\paragraph{Energy Gain ($G_e$)}
The agent is incentivized to maximize the kinetic and elastic potential energy of the leg at any given time step. The reasoning behind this term is an approximation of the leg by spring with mean length $x_o$. This reward term promotes an oscillatory behaviour leading to high enough velocities for hopping. The corresponding term is calculated as:
\begin{equation}
	G_e = \dot{x}^2 + (x-x_{o})^2
\end{equation}
where $x$ and $\dot{x}$ are the base height and velocity, respectively. $x_o$ is the base height for the initial standing position of the leg.
\paragraph{Height barrier penalty ($P_{h}$)}
The agent is penalized exponentially when the base height crosses the desired height command $x^d$.
\begin{equation}
	P_h = \begin{cases} 
		1 - e^{x - x^d}, & \text{if} \: x \ge x^d \\
		0, & \text{otherwise}
	\end{cases}
\end{equation}

\paragraph{Jerky Action Penalty ($P_j$)}
Sudden changes in the output torques can cause shakiness in the hardware, making the policy hard to transfer. Therefore, the agent is penalized for large differences in consecutive actions.
\begin{equation}
	P_j = \sum_{i=0}^{2}(a_t^i - a_{t-1}^i)^2
\end{equation}
\paragraph{Joint constraints penalty ($P_{jp}, P_{jv}$)}
It is desired to keep the joint position limits within some pre-defined constraints to avoid self-collisions and prevent arbitrary configurations. The joint velocities should be reasonably bounded for successful sim-to-real transfer. These constraints are imposed with negative penalties. The penalty for the position limit is structured such that it becomes significant around the limits and beyond them but stays reasonably low elsewhere. It is calculated as:
\begin{equation}
	P_{jp} = \sum_{i=0}^{2} \begin{cases} 
		e^{-10(q_i - q_i^l)} + e^{10(q_i - q_i^h)}, & \text{if} \: q_i^l \le q_i \le q_i^h \\
		1, & \text{otherwise}
	\end{cases}
\end{equation}
Here, $q^l_i$ and $q^h_i$ denote the lower and upper joint limits, respectively. To reasonably constrain the search space for the agent, we used a PD controller to bring the joints back within bounds if joint limits are passed during training. The joint velocities are penalized if they cross the saturation limits for the motors.
\begin{equation}
	P_{jv} = \sum_{i=0}^{2} \begin{cases} 
		0, & \text{if} \: -\dot{q}_i^h \le \dot{q_i} \le \dot{q}_i^h \\
		\dot{q_i}^2 - \dot{q}_i^{h^2}, & \text{otherwise}
	\end{cases}
\end{equation}
Here, $\dot{q}_i^h$ is the maximum desired joint velocity.

The final expected reward is calculated as:
\begin{equation}
	R = w_1G_e - w_2P_h - w_3P_j - w_4P_{jp} - w_5P_{jv}
\end{equation}
The weights used during training are $w_1 = 0.5$, $w_2 = 2$, $w_3 = 0.05$, $w_4 = 0.02$, and $w_5 = 0.005$.

\subsection{Simulation to Reality Transfer}
We use a custom gym \cite{gym} environment with MuJoCo physics engine \cite{6386109} for training in simulation. As explored in \cite{https://doi.org/10.48550/arxiv.2103.04616}, MuJoCo is well suited for robotics and reinforcement learning problems as it provides a wide range of solver parameters and settings, which can be adapted and optimized for many use cases. The policies trained in MuJoCo with default model and simulation parameters failed to transfer to the hardware. Therefore, we  optimised the simulation parameters to narrow the sim-to-real gap.

\subsubsection{Simulation Parameter Optimisation}
The goal of this step is to match 
simulation dynamics to the real robot using simple training trajectories.

\begin{figure}[t]
	\centering
	\includegraphics[width=0.8\linewidth]{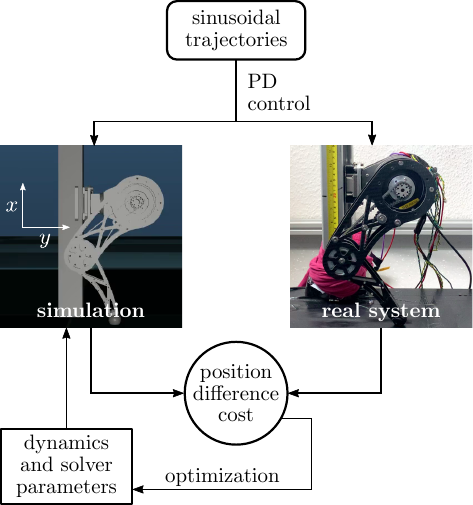}  
	\caption{Parameter optimization pipeline.}
	\label{fig:sys_id_pipeline}
\end{figure}

\paragraph{Trajectory Generation and Data Collection}
A varied set of task-space, hand-tuned sinusoidal trajectories is generated for two different system configurations. These include a fixed-base configuration, where the leg is suspended in the air, and a moving-base configuration, where the leg 
comes in contact with the ground. For the fixed-base configuration, the template trajectories are:
\begin{align}
\begin{split}
	x & = (A\cos\left(\frac{2\pi t}{T_1}\right) + \epsilon)\cos(\theta) \\
	y & = (A\cos\left(\frac{2\pi t}{T_1}\right) + \epsilon)\sin(\theta) \\
	\theta & = -\frac{\pi}{2}\cos\left(\frac{2\pi t}{T_2}\right)	
\end{split}
\end{align}
Here, $\epsilon = L_1 + L_2 - A$ is the trajectory offset from the origin, given $L_1$ and $L_2$ are the shank and calf link lengths for the leg. The periods ($T_1, T_2$) and the amplitude $A$ are varied such that the maximum workspace of the leg is covered.
The moving-base trajectory only consists of vertical trajectories with a few segments fast enough to break ground contact. These trajectories closely imitate the hopping configuration of the leg and are given as:
\begin{align}
\begin{split}
	x & = A\cos\left(\frac{2\pi t}{T_1}\right) + \epsilon \\
	y & = 0
\end{split}
\end{align}
The joint-level trajectories are obtained through inverse kinematics and tracked on the hardware with a PD controller running at a frequency of $200Hz$. The controller gains are fixed and the target velocity set to 0. For both fixed-base and floating base 
configuration, 240 s of data was recorded where $T_1 \in \{0.75, 0.5, 0.25\}$, 
$T_2 \in \{10, 20\}$, and $A \in \{0.15, 0.1, 0.05\}$. 
Joint positions, velocities, and resulting motor torques are 
recorded on the actual hardware.
\paragraph{Simulation Parameters Optimisation}
Using the hardware trajectories, we optimize for the simulation's dynamics and solver parameters. We use the same PD controller running at the same frequency to track the generated trajectories in simulation, with gains adjusted for the motors' internal gear ratio. We optimize for the following set of simulation parameters:
\begin{itemize}
 \item \emph{Dynamic Parameters:} The simulation model's dynamic parameters to be optimized involve the friction loss, damping and armature (rotor inertia) values for the hip and knee motors, friction loss and damping for the rail, which is modeled by a passive prismatic joint, and the link inertias.
 \item \emph{Solver Parameters:} Time constant and damping ratio are two of the solver parameters, characteristic of the mass-spring-damper constraint modeling of MuJoCo. These parameters are optimized to modulate the contact model between the leg and the plane.
\end{itemize}
CMA-ES \cite{6790628} is used to optimize these parameters with a cost 
on the cumulative joint position difference between simulation trajectories and recorded real hardware data at each time step.
\begin{equation}
	J(q) = \sum_{t = 0}^{t_f} \sum_{i = 0}^{1}(q_{i_{\mathrm{sim}}}^t - q_{i_{\mathrm{real}}}^t)^2
\end{equation}
In high dimensional optimization problems, such as this, it can become hard for the solver to converge and find an optimal solution. Dynamic coupling also occurs between the parameters, 
potentially leading to low cost but poor transfer to the hardware. To prevent these issues, the parameters that can be roughly estimated during modeling, i.e. link inertia and solver parameters, 
are fixed in a first optimization pass. In a second pass, we optimize all dynamic and solver 
parameters while placing bounds on the friction, damping, and armature parameters derived from the first pass. This 
helps converging to an optimal and pragmatic solution. Fig. \ref{fig:sys_id_pipeline} visualizes the optimization procedure.

\subsubsection{Policy Robustness}
Two methods are employed during training to make the policy robust to delays, noise, and other disturbances.

\paragraph{Delays}
As mentioned before, the observation space consists of sensor readings from the last three consecutive time steps. In addition, the observation data over the last ten time-steps is stored in a buffer. While training, with a probability of 0.5 at each time step, data of three time steps is randomly sampled from the buffer in 
correct temporal sequence and used as observation instead. This helps to simulate plausible delays on the real system, effectively making the policy more robust.

\paragraph{Noise}
Noise is added to the joint data and the torques given by the policy to simulate 
sensor noise and control inaccuracies. At each time step, the noise is sampled from a uniform distribution ranging from $-\lambda u$ to $\lambda u$, where $\lambda$ is the error range and $u$ is the observed value. We set $\lambda = 0.05$ for the joint positions and velocities, and $\lambda = 0.15$ for the output torques.

\section{Results}
\label{sec_results_discussions}
{\renewcommand{\arraystretch}{1.3}
\begin{table}[t]
\caption{The simulation parameters obtained after CMA-ES optimization.}
\label{tab:parameters}
\begin{center}
	\begin{tabular}{ |c|c|c|c| } 
		\hline
		\textbf{Joint} & Friction loss & Damping & Armature \\
		\hline
		Rail Prismatic Joint & $0.7024$ & $1.0724$ & -\\
		\hline 
		Hip Joint & $0.4364$ & $0.0005$ & $0.00004$ \\
		\hline 
		Knee Joint & $0.0015$ & $0.1441$ & $0.0001$ \\ 
		\hline
	\end{tabular} \\
	\vspace{5 mm}
	\begin{tabular}{ |c|c| } 
	\hline
	\textbf{Parameter} & Value \\
	\hline
	Hip Link Z Inertia & $0.004061$ \\
	\hline 
	Knee Link Z Inertia & $0.000845$  \\
	\hline 
	Time Constant & $0.0911$  \\ 
	\hline
	Damping Ratio & $0.6678$ \\
	\hline
	\end{tabular}
\end{center}
\end{table}

The optimization of simulation parameters described in Section~\ref{sec:mats_and_meths} lead to the parameters shown in 
Table~\ref{tab:parameters}.
Training for the jump heights $0.25$ m, $0.3$ m, and $0.35$ m in simulation 
yielded a controller that is able to interpolate between these three 
desired jump heights, 
showing that the controller learned an approximation of the 
task space inverse dynamics for this problem. Figure~\ref{fig:mix_30s_interpol} shows the jump 
height of a $30$ s trial with an initial desired jump height of $0.25$ m. 
Every $5$ s the desired jump height is increased by $0.02$ m. While there is a significant deviation of the actual average jump height especially for intermediate commands, the mapping from desired to actual jump heights is monotonic. 

\begin{figure}[ht]
	\centering
	\includegraphics[width=.85\linewidth]{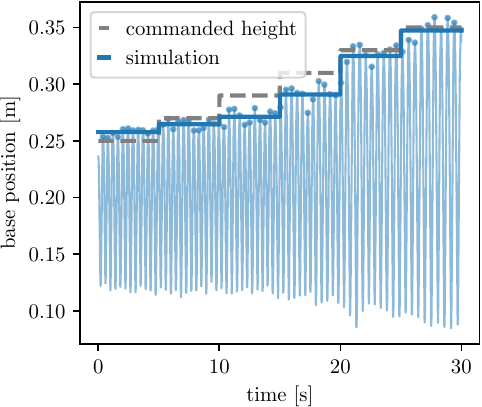}  
	\caption{Jumping heights of the simulated robot in a 30 second trial 
		with 
		monotonically increasing desired jump heights. The desired 
		heights increase from $0.25$ m to 
		$0.35$ m in increments of $0.02$ m. The controller was trained with only three desired jump heights of $0.25$ m, $0.30$ m, and $0.35$ m.}
	\label{fig:mix_30s_interpol}
\end{figure}

\begin{figure}[ht]
	\centering
	\includegraphics[width=.8\linewidth]{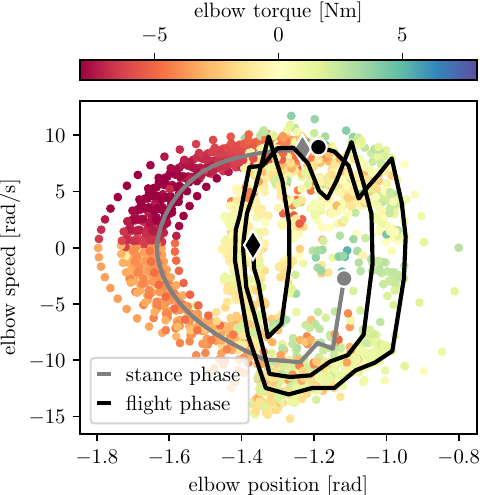}  
	\caption{Controller torque output during $10$ s of continuous 
		jumping with a commanded height of $0.30$ m in phase space of 
		the elbow joint. The black (grey) line is an example phase space 
		trajectory for one flight (stance) phase. Circles and diamonds denote starting and end 
		points of the jump phase. 
	}
	\label{fig:controller_output_stance_flight}
\vspace{-0.8cm}
\end{figure}

To assess the implicit contact detection of the controller, we analyse the 
applied torque in simulation to the elbow joint for a $10$ s trial 
with a commanded jump height of $0.30$ m. Figure \ref{fig:controller_output_stance_flight} shows the controller torque 
output for all encountered configurations in the phase space of the 
actuated elbow joint. It is evident that during the stance phase, the 
controller applies significantly higher torques than in the flight 
phase, to generate the lift-off. In addition, the control torque 
increases after the minimal attitude is reached. This strongly implies that 
the controller indeed detects ground contact, solely based on the proprioceptive 
observation of joint positions and velocities.

The controller trained in simulation was tested on the 
real robot without further adjustment. Figure~\ref{fig:mix_15_s_real_sim} 
shows the base height trajectories for simulated and real robot for 
a trial with changing desired jump height. Whereas the real jump height 
is lower than the commanded height, the ordering of jump heights is as 
intended, i.e. a higher desired height leads to a higher actual jump height. 
The offset between commanded and actual jump height lies between 
$0.04$ m and $0.06$ m. 
We observe that the temporal structure of 
consecutive jumps differs between simulation and the real robot. 
For a desired jump height of $0.25$ m, the real robot shows jump 
heights alternating between $\approx 0.2$ and $\approx 0.24$ m. 
For a commanded height of $0.35$ m, the jump frequency on the real 
robot is slightly reduced, because two jumps, around 11 and 12.5 s, were 'skipped'. Note that the absolute values of the jump heights on 
the real system are not exact, since it is determined by 
tracking the center of the upper motor in video recordings. This 
induces some noise on the measurement. In addition, a height dependent 
small parallaxis error can be expected.

\begin{figure}[!th]
	\centering
	\includegraphics[width=.85\linewidth]{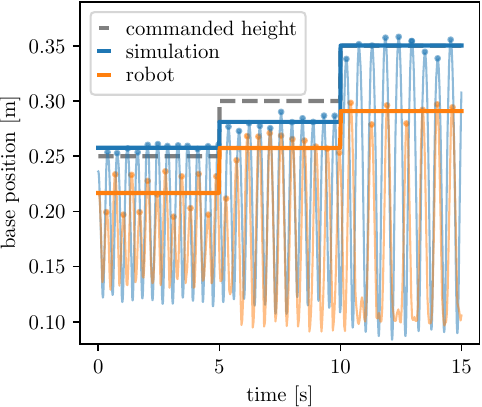}  
	\caption{Jumping heights of simulated and real robot for 15 seconds. The commanded 
	jump height is 0.25 m for the first 5 s,  0.30 m for the next 5 s, 0.35 m for the last five seconds. Both in simulation and the real robot, 
	increasing the desired jump height leads to higher actual jumps. 
}
	\label{fig:mix_15_s_real_sim}
\end{figure}

\begin{figure}[!th]
	\centering
	\includegraphics[width=0.85\linewidth]{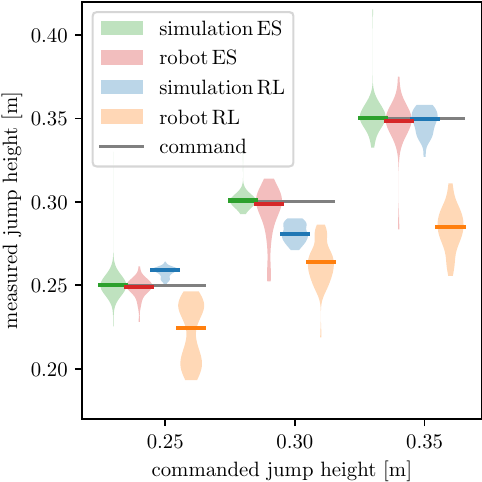}
	\caption{Jump height distributions for different commands, in simulation and on 
		the real system. Colored lines show the medians of the 
		distributions, gray lines the respective commanded height. 
		In simulation, the distributions for the RL controller are completely 
		non-overlapping, thus clearly significantly different. On the real 
		system, the distributions show some overlap. 
        For comparison, the distributions of jump heights generated by an ES controller are also shown.
	}
	\label{fig:jump_height_distribution}
	\vspace{-0.6cm}
\end{figure}

For a statistical analysis of the jump height distribution for 
varying height commands, the data of the 15 s trial shown in 
Figure~\ref{fig:mix_15_s_real_sim} is merged with 15 s trials of 
fixed jump heights at $0.25$ m, $0.30$ m, and $0.35$ m, both in simulation 
and the real robot (also see the accompanying video available as the multimedia attachment). Figure~\ref{fig:jump_height_distribution} show the 
resulting jump height distributions for the trained controller, along
with simulation results for an energy shaping controller as reference. 
While the distributions of RL based jumping are 
completely separated for the simulation, the real system tests 
show overlaps between neighbouring jump height distributions. 
This can at least partly be attributed to the noise induced by the pixel tracking used to estimate the heights in the real experiments. The bimodal nature of the distribution for a commanded height of $0.25$ m is a consequence of 
the alternating jump heights, also seen in Figure \ref{fig:mix_15_s_real_sim}. We use a Wilcoxon-Ranksum test to evaluate the difference between neighbouring distributions. Both in simulation and on the real system, all neighbouring distributions are significantly different at $p < 0.001$. 

Our baseline energy shaping controller worked remarkably well in both simulation and on real hardware (see Fig.~\ref{fig:jump_height_distribution}). This is as expected since we exploit the model knowledge and physics that captures the essence of jumping task. However, it requires expert knowledge to tune the contact detection threshold and other controller gains which can be time consuming. Our proposed End-to-End RL controller does not require such expert knowledge and demonstrates a similar trend for different jumping heights. The standard deviation in jumping height is even smaller in some cases, especially in simulation. However, there is substantial room for improvement in the performance of the RL controller on the real system in comparison to the baseline ES controller.

\section{Discussion}
\label{sec_discussion}
The main objective was to find a jumping 
controller mapping proprioceptive feedback to 
torque control, including the avoidance of height estimation and 
PD control strategies as used by~\cite{bogdanovic2021model}. 
Thus it may seem counterintuitive that we impose soft joint limits with a PD controller
and use the base height for 
reward calculation. However, since 
variable height jumping cannot be defined without the notion 
of height, it is strictly necessary information 
for the agent such that the task 
space inverse dynamics can be approximated. We want to emphasize 
though that the height is only used in the reward during training, and 
is not required as direct feedback to the controller. The 
soft joint limits serve as a gentle exploration guiding strategy, similar 
to initial example trajectories as used by~\cite{bogdanovic2021model}. 
On the hardware, they are still in place for safety reasons, but are rarely crossed.
Thus, our control approach can be considered truly end-to-end.
In the following, further 
features and critical design decisions are discussed in more depth.

A prerequisite for successful transfer to the real system is a 
small simulation to reality gap. Prior to developing the 
current approach, the more common technique of domain 
randomization~\cite{peng2018sim} was also tested, which generated unsatisfying 
behavior transfer. Our approach is adapted from \cite{kaspar2020sim2real}, 
who used simulation parameter optimization to make trajectories in 
simulation follow real recorded training data.
We extend this approach by 
introducing a two stage process for high dimensional parameter spaces 
and showing the applicability to collision rich and dynamic tasks. 
We chose to keep inertia parameters fixed in the first stage, 
since they can be reasonably well estimated from the structure, 
whereas other dynamical parameters are much harder to infer a priori. 
We also noted that identifying the rotor inertia was 
crucial. While not necessary for less dynamic behavior such as
walking, rotor inertia becomes more influential for highly dynamic 
motions. 
The superior simulation to reality transfer can be 
explained by domain randomization leading to a trade off 
between generality over a range of parametrizations to optimality 
on the actual hardware, which has well defined parameters. In contrast, 
the method we propose is more akin to dynamic system identification. However, the
target is not the true physical parameters, but the closest possible representation 
of the system dynamics within the simulation.

To make the policy robust to expectable delays on the real system, we used 
random sampling from an observation buffer. 
This random sampling technique is easy to implement without having to 
know the 
exact delays and their distributions. 
An alternative approach would be to use 
an actuator model as suggested by \cite{hwangbo2019learning} to learn 
quadruped walking. In their case, 
a good motor model was probably more relevant since the robot's legs 
use series elastic actuators, 
which are expected to have more complex delay dynamics. If this is not the 
case, we argue for our method as a simpler solution.

The policy shows good interpolation performance for 
height values that were not explicitly included in the training. This 
suggests that the policy implicitly learned a task space inverse dynamics model of the 
system. This assumption is further supported by the implicit detection of 
different jump phases. 
However, height tracking shows relatively higher 
deviations at intermediate commands around 0.3 m. This could be an 
issue of the neural network not having enough capacity to represent the full
dynamics. A thorough hyperparameter tuning of the network architecture could 
improve the results, but is out of scope for this paper.

The remaining differences in the jump heights between simulation and 
reality
can be a consequence of non-optimal dynamic parameters of the simulator. 
However, we noted that adding more data to the parameter optimization 
pipeline did not 
significantly change the optimization result. Another explanation could be additional,  
unmodelled non-linear dynamics such as motor backlash, motor torque saturation, or state dependent 
sensor noise. A strategy to improve performance  
without having to explicitly model these effects is to continue training the 
controller on the real system directly, using the current policy as a starting point. 
For this, the used SAC algorithm is particularly well suited  \cite{haarnoja2018learning}.

\section{Conclusion}
\label{sec_conclusion}
In summary, we presented a method 
to train a unified torque controller for continuous hopping 
with a monoped robot. The controller is able to interpolate 
between jump heights and implicitly detect relevant jump phases and act 
accordingly. The simulation to reality mapping procedure 
eliminates the need of parameter tuning for behavior transfer. 
The trained policy realizes a direct mapping from proprioceptive feedback 
to torque control.
To the authors' knowledge, this is the first reported 
end-to-end training procedure for a jump height adjustable monoped torque controller. However, much needs to be done to bring the height tracking accuracy of this approach closer to the model based energy shaping control.
Future research directions  
include a thorough hyperparameter tuning of the neural network architecture 
to improve the jump height interpolation in simulation, as well as continued 
training on the real system to mitigate the effect of residual dynamics modeling inaccuracies of the simulator. 
We also plan to integrate this work in the RealAIGym ecosystem~\cite{2022_rss_realaigym} similar to other canonical underactuated systems like simple pendulum~\cite{Wiebe2022}, double pendulum~\cite{2023_ram_wiebe_double_pendulum}, and AcroMonk~\cite{2023_mahdi_acromonk}.

\bibliographystyle{IEEEtran}
	\bibliography{references}

\end{document}